\documentclass{sig-alternate-05-2015}

\usepackage{CJK,algorithm,algorithmic,amssymb,amsmath,array,epsfig,graphics,multirow,array,float,subfigure,verbatim,epstopdf}
\usepackage{acmcopyright}
\usepackage{enumitem}
\usepackage{times}
\usepackage{url}
\usepackage{latexsym}
\usepackage{color}
\usepackage{hhline}
\definecolor{Orange}{rgb}{1,0.5,0}

\DeclareMathOperator*{\argmax}{\arg\!\max}

\begin{document}

\setcopyright{acmcopyright}





%

\title{Building a Fine-Grained Entity Typing System Overnight for a New X (X = Language, Domain, Genre)}

%
%
%
%
%

\numberofauthors{4} 
%

%
%
\author{
\alignauthor
Lifu Huang\\
       \affaddr{Rensselaer Polytechnic Institute}\\
       \affaddr{Troy, NY, USA}\\
       \email{huangl7@rpi.edu}
\alignauthor
Jonathan May\\
       \affaddr{Information Sciences Institute}\\
       \affaddr{Marina del Rey, CA, USA}\\
       \email{jonmay@isi.edu}  
\alignauthor
Xiaoman Pan\\
       \affaddr{Rensselaer Polytechnic Institute}\\
       \affaddr{Troy, NY, USA}\\
       \email{panx2@rpi.edu}    
\and
\alignauthor
Heng Ji\\
       \affaddr{Rensselaer Polytechnic Institute}\\
       \affaddr{Troy, NY, USA}\\
       \email{jih@rpi.edu} 
}       

\maketitle
\begin{abstract}

Recent research has shown great progress on fine-grained entity typing. Most existing methods require pre-defining a set of types and training a multi-class classifier from a large labeled data set based on multi-level linguistic features. They are thus limited to certain domains, genres and  languages.  In this paper, we propose a novel unsupervised entity typing framework by combining symbolic and distributional semantics. We start from learning general embeddings for each entity mention, compose the embeddings of specific contexts using linguistic structures, link the mention to knowledge bases and learn its related knowledge representations. Then we develop a novel joint hierarchical clustering and linking algorithm to type all mentions using these representations. This framework doesn't rely on any annotated data, predefined typing schema, or hand-crafted features, therefore it can be quickly adapted to a new domain, genre and language. Furthermore, it has great flexibility at incorporating linguistic structures (e.g., Abstract Meaning Representation (AMR), dependency relations) to improve specific context representation. Experiments on genres (news and discussion forum) show comparable performance with state-of-the-art supervised typing systems trained from a large amount of labeled data. Results on various languages (English, Chinese, Japanese, Hausa, and Yoruba) and domains (general and biomedical) demonstrate the portability of our framework.

\end{abstract}
\section{Introduction}
\label{sec:introduction}

Entity typing is an important and challenging task which aims to automatically assign types to entity mentions in unstructured documents. Previous entity typing work mainly focused on a small set of predefined types. For example,  MUC-7~\cite{Grishman1996}  defined the three most common types: Person, Organization, and Location. ACE\footnote{http://www.itl.nist.gov/iad/mig/tests/ace/} separated Geo-Political Entities from (natural) Locations, and introduced Weapons, Vehicles and Facilities. These entity types are very useful for many downstream Natural Language Processing (NLP) and Information Retrieval (IR) tasks such as entity disambiguation~\cite{cucerzan2007,bunescu2006,han2009}, entity linking~\cite{rodriguez2003determining,lin2012entity,durrett2014joint}, relation extraction~\cite{hasegawa2004discovering,culotta2004dependency,bunescu2005shortest}, knowledge base completion~\cite{ji2011knowledge,dredze2010entity,suchanek2007yago}, question answering~\cite{molla2006named,sun2015open}, and query analysis~\cite{balog2012hierarchical,bazzanella2010searching}. 

Recent work~\cite{Ling12fine-grainedentity,lee2006fine} suggests that using a larger set of fine-grained types can lead to substantial improvement for these downstream NLP applications. 
Ling et al. \cite{Ling12fine-grainedentity} described three remaining challenges for the fine-grained entity typing task: selection of a set of fine-grained types, creation of annotated data and linguistic features, and disambiguation of fine-grained types. 
 Most existing approaches~\cite{fleischman2002fine,giuliano2009fine,ekbal2010assessing,Ling12fine-grainedentity,yosef2012hyena,nakashole2013fine,GillickLGKH14,yogatama2015embedding,Corro2015} addressed these problems in a ``distant supervision'' fashion, by pre-defining a set of fine-grained types based on existing knowledge bases (KBs) (e.g., DBPedia, Freebase, YAGO), and creating annotated data based on anchor links in Wikipedia articles or types in KBs, then training a multi-class classifier to predict types. 
 However, the training data acquired in this way often includes a lot of noise, because the entity types are assigned without considering specific contexts. For example, in DBPedia, ``\emph{Mitt Romney}'' is assigned with a set of types including ``\emph{President},'' ``\emph{Scholar},'' ``\emph{Teacher},'' ``\emph{Writer}'' and so on. More importantly, 
 these types are limited to a pre-defined set, and may not be easily adapted to a new domain, genre, or language. 

In this work, we propose an unsupervised fine-grained entity typing framework. 
Let's take a look at the following motivating examples: 


\begin{enumerate}[label={E}\arabic*.]

\item \textbf{Mitt Romney} was born on March 12, 1947, at \textbf{Harper University Hospital} in \textbf{Detroit}, \textbf{Michigan}, the youngest child of automobile executive \textbf{George Romney}.

\item \textbf{Yuri dolgoruky}, \underline{\emph{equipped}} by  Bulava nuclear-armed missile, is the first in a series of new \underline{\emph{nuclear submarines}} to be \underline{\emph{commissioned}} this year. 

\item \textbf{OWS} \underline{\emph{activists}} were part of \underline{\emph{protest}}.


\item The effects of the \textbf{MEK} inhibitor on total \textbf{HER2} , \textbf{HER3} and on phosphorylated \textbf{pHER3} were dose dependent.

\end{enumerate}

In E1, the mentions like ``\emph{Mitt Romney},'' ``\emph{George Romney},'' ``\emph{Detroit},'' and ``\emph{Michigan}'' are commonly used and have no type ambiguity. Thus, their types can be easily determined by their general semantics. 
Our first intuition is that: 

\vspace{+0.2cm}
\textbf{Heuristic 1: } \textbf{\emph{The types of common entities can be effectively captured by their general semantics.}} 
\vspace{+0.2cm}

\begin{table}[!htb]
\centering
\begin{tabular}{c|c|c|c}
\hline
\multicolumn{2}{c|}{\textbf{Yuri Dolgoruky}} & \multicolumn{2}{c}{\textbf{OWS}}
\\ \hline
Vsevolodovich & 0.828 & nks & 0.595 \\ \hline
Yurevich & 0.803 & sh$\_$* & 0.585 \\ \hline
Romanovich & 0.801 & cked & 0.581 \\ \hline
Pronsk & 0.801 & issa & 0.580 \\ \hline
Dolgorukov & 0.798 & sh & 0.0.580 \\ \hline 
Yaroslavich & 0.796 & cks & 0.577 \\ \hline
Feodorovich & 0.793 & **** & 0.573 \\ \hline
Shuisky & 0.784 & ckers & 0.567 \\ \hline
Pereyaslavl & 0.784 & $\#\_$cking & 0.566 \\ \hline
Sviatoslavich & 0.781 & ked & 0.565 \\ \hline
\end{tabular}
\caption{Top-10 Most Similar Concepts and Scores based on General Embeddings Learned from Wikipedia Articles}
\label{similarConcepts}
\vspace{-0.2cm}
\end{table}

However, many entities are polysemantic and can be used to refer to different types in specific contexts. For example,
``\emph{Yuri Dolgoruky}'' in E2, which generally refers to the Russian prince, is the name of a submarine in this specific context. Likewise, ``\emph{OWS}'' in E3, which refers to ``\emph{Occupy Wall Street},'' is a very novel emerging entity. It may not exist in the word vocabulary, and its general semantics may not be learned adequately due to its low frequency in the data. Table~\ref{similarConcepts} shows the top-10 most similar entities to ``\emph{Yuri Dolgoruky}'' and ``\emph{OWS}'' based on word embeddings learned from Wikipedia articles. Such types are difficult to capture with general semantics alone, but can be inferred by their specific contexts, like ``\emph{nuclear submarines},'' ``\emph{equip},'' ``\emph{commission},'' ``\emph{activists},'' and ``\emph{protest}''. Thus, our second intuition is that: 

\vspace{+0.2cm}
\textbf{Heuristic 2: }\textbf{\emph{The types of uncommon, novel, emerging, and polysemantic entities can be inferred by their specific contexts.}}
\vspace{+0.2cm}

In E4, ``\emph{MEK},'' ``\emph{HER2},'' ``\emph{HER3},'' and ``\emph{pHER3}'' are biomedical domain specific entities. Their types can be inferred from domain-specific KBs. For example, the properties for ``\emph{pHER3}'' in biomedical ontologies include ``\emph{Medical},'' ``\emph{Oncogene},'' and ``\emph{Gene}.'' Therefore, we derive the third intuition:

\vspace{+0.2cm}
\textbf{Heuristic 3: }\textbf{\emph{The types of domain specific entities largely depend on domain-specific knowledge.}}
\vspace{+0.2cm}




Based on these observations and intuitions, we take a fresh look at this task. For the first time, we propose an unsupervised fine-grained entity typing framework that combines general entity semantics, specific contexts, and domain specific knowledge. Without the needs of predefined typing schema, manual annotations and hand-crafted linguistic features, this framework can be easily applied  to new domains, genres, or languages. 
The types of all entity mentions are automatically discovered based on a set of clusters, which can capture fine-grained types customized for any input corpus. We compare the performance of our approach with state-of-the-art name tagging and fine-grained entity typing methods, and show the performance on various domains, genres, and languages. 

The novel contributions of this paper are as follows:

\begin{enumerate}[label=$\bullet$]
\item We present an unsupervised entity typing framework that 
requires no pre-defined types or annotated data. Instead, it can automatically discover a set of fine-grained types customized for the input corpus.

\item We combine three kinds of representations, which are language-, genre-, and domain-independent and very effective, to infer the types of entity mentions. 

\item We design an unsupervised entity linking-based type naming approach to automatically generate fine-grained type labels, and jointly optimize clustering and linking.

\end{enumerate}
\section{Related Work} 


Several recent studies have focused on fine-grained entity typing. Fleischman et al. \cite{fleischman2002fine} classified person entities into eight fine-grained subtypes based on local contexts. Sekine \cite{Sekine2008} defined more than 200 types of entities. The Abstract Meaning Representation (AMR)~\cite{Banarescu2013} defined more than 100 types of entities. FIGER~\cite{Ling12fine-grainedentity} derived 112 entity types from Freebase~\cite{bollacker2008freebase} and trained a linear-chain CRF model~\cite{lafferty2001conditional} for joint entity identification and typing. 
Gillick et al. \cite{GillickLGKH14} and Yogatama et al. \cite{yogatama2015embedding} proposed the task of context-dependent fine-grained entity typing, where the acceptable type labels are limited to only those deducible from local contexts (e.g., a sentence or a document). Similar to FIGER, this work also derived the label set from Freebase and generated the training data automatically from entities resolved in Wikipedia. 
Lin et al. \cite{lin2012no} proposed propagating the types from linkable entities to unlinkable noun phrases based on a set of features. HYENA ~\cite{yosef2012hyena} derived a very fine-grained type taxonomy from YAGO~\cite{hoffart2013yago2} based on a mapping between Wikipedia categories and WordNet synsets. 
This type structure incorporated a large hierarchy of 505 types organized under 5 top level classes (person, location, organization, event, artifact), with 100 descendant types under each of them. 
Although these methods can handle multi-class multi-label assignment, the automatically acquired training data is often too noisy to achieve good performance. 
In addition, the features they exploited are language-dependent, and their type sets are rather static.

Our work is also related to embedding techniques. Turian et al. \cite{turian2010word} explored several unsupervised word representations including distributional representations and clustering-based word representations. 
Mikolov et al. \cite{mikolov2013linguistic} examined vector-space word representations with a continuous space language model. 
Besides word embedding, several phrase embedding techniques have also been proposed. Yin et al. \cite{yin2014exploration} computed embeddings for generalized phrases, including both conventional linguistic phrases and skip-bigrams. Mitchell et al. \cite{MitchellLapata2010} proposed an additive model and a multiplicative model.  Linguistic structures have been proven useful to capture the semantics of basic language units~\cite{socher2013recursive,hermann2013role,Socher2013,BowmanPM14}. Socher et al. \cite{Socher2013} designed a DT-RNN model to map sentences into compositional vector representations based on dependency trees. Hermann et al. \cite{hermann2013role} explored  a novel class of Combinatory Categorial Autoencoders to utilize the role of syntax in Combinatory Categorial Grammar to model compositional semantics. Socher et al. \cite{socher2013recursive} designed a Recursive Neural Tensor Network (RNTN) to compute sentiment compositionality based on the Sentiment Treebank. Bowman et al. \cite{BowmanPM14} designed three sets of experiments to encode semantic inference based on compositional semantic representations. Compared to these efforts, in this work, we attempt to compose the context information to infer the fine-grained types. Considering not all contexts are meaningful, we carefully selected specific types of relations (AMR relations or dependency relations) to capture concept-specific local contexts instead of sentence-level or corpus-level contexts. 

\section{Approach Overview}





\begin{figure*}[!htp]
\centering
\includegraphics[width=.9\textwidth]{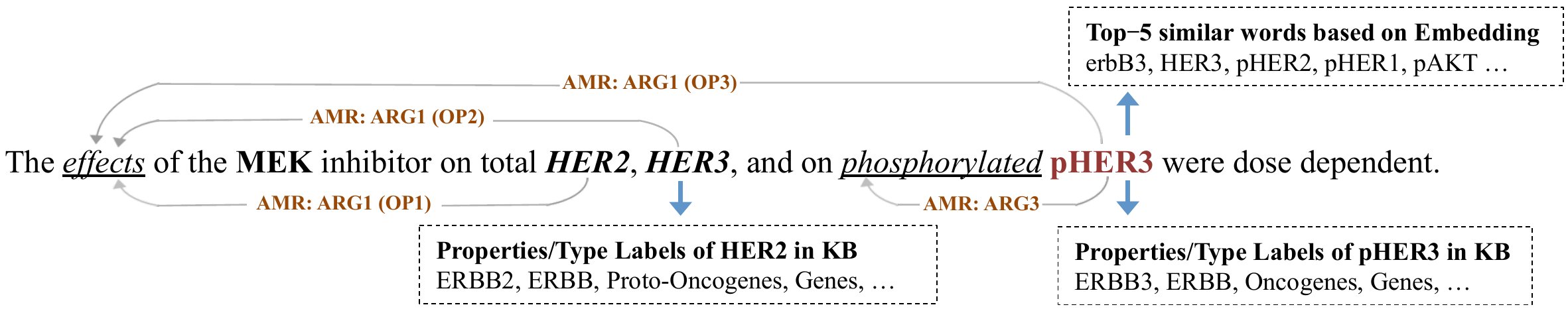}
\caption{Information that can be Used to Infer the Type of ``pHER3''}
\label{walkThrough}
\end{figure*}

\begin{figure}[!htp]
\centering
\includegraphics[width=.43\textwidth]{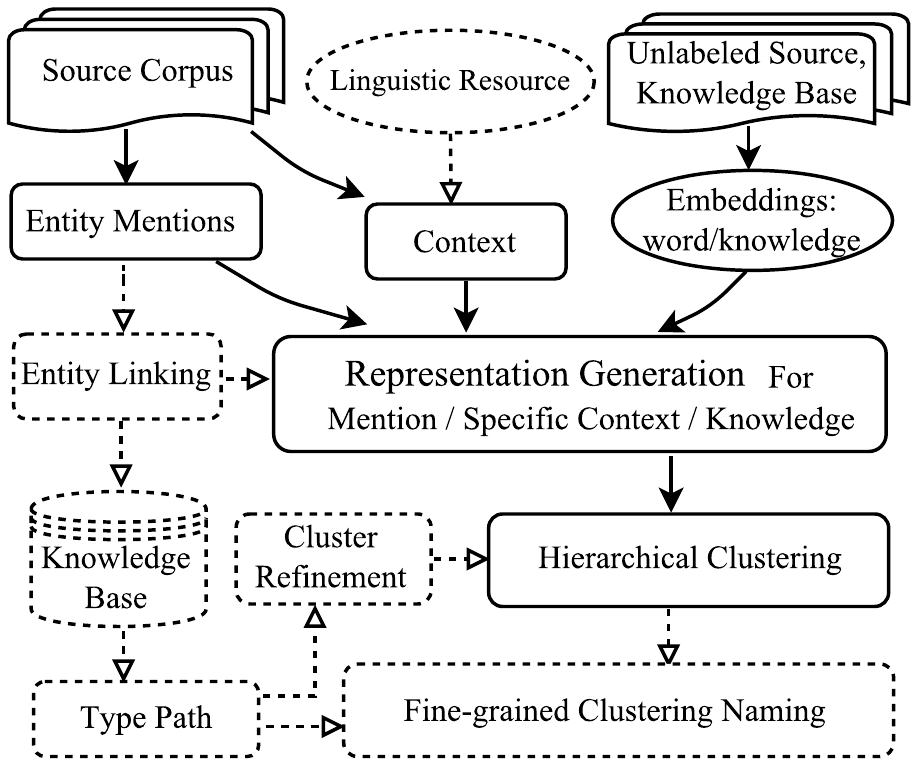}
\caption{Approach Overview (The solid boxes are required while the dotted boxes are optional)}
\label{workflow}
\vspace{-0.1cm}
\end{figure}

Figure~\ref{workflow} illustrates the overall framework of our system, which takes the boundaries of all entity mentions as input and produces a type label for each mention as output. The framework starts from learning three kinds of representations: 
\begin{enumerate}[label=(\roman*)]
\item a general entity distributed representation based on global contexts; 
\item a specific context representation, to model local context words based on linguistic structures.
\item a knowledge representation, to model domain-specific knowledge for each mention. 
\end{enumerate}

For example, Figure~\ref{walkThrough} shows these three representations that can be used to infer the type of ``\emph{pHER3}''. After learning general and context-specific embeddings, we apply unsupervised entity linking to link entity mentions to a domain-specific knowledge base. Based on the linking results, we can determine the knowledge representation and extract a type path for each highly linkable entity mention. Finally, we use these three representations as input to a hierarchical X-means clustering algorithm~\cite{Pelleg2000} to discover the hierarchical type structure of each mention. At each layer of the hierarchy, we also design an optimal partition search algorithm to conduct feature voting and obtain the optimal clustering and typing results.

\section{Representation Generation}
\subsection{General Entity Representation}

Based on Heuristic 1, we can infer the types of most entity mentions. 
For example, ``\emph{Mitt Romney}'' and ``\emph{John McCain}'' are both politicians from the same country, ``\emph{HER2}'' and ``\emph{HER3}'' refer to similar ``\emph{ERBB (Receptor Tyrosine-Protein Kinase)}'' and thus they have the same entity type ``\emph{Enzyme}''. We start by capturing the semantic information of entity mentions based on general lexical embedding, which is an effective technique to capture general semantics of words or phrases based on their global contexts. Several models~\cite{mikolov2013distributed,mikolov2013efficient,zhao2014learning,MitchellLapata2010} have been proposed to generate word embeddings. 
Here, we utilize the Continuous Skip-gram model~\cite{mikolov2013efficient} based on a large amount of unlabeled in-domain data set.

Most entity mentions are multi-word information units. In order to generate their phrase embeddings, we compared two methods: (1) the model proposed by Yin et al. \cite{yin2014exploration}, which learned phrase embeddings directly by considering a phrase as a basic language unit, and (2) a simple element-based additive model ($z = x_1 + x_2 + ... + x_i$)~\cite{mikolov2013efficient}, where $z$ represents a phrase embedding and  $x_1, x_2, ..., x_i$ represent the individual embeddings of the words in $z$. We found the former performed better because most entity mentions appear more than 5,000 times in the large in-domain data set, and thus the embedding of phrases can be learned efficiently.

\subsection{Specific-Context Representation}

General embeddings can effectively capture the semantic types of most entity mentions, but many entities are polysemantic and can refer to different types in specific contexts. For example, ``\emph{ADH}" in the biomedical domain can be used to refer to an enzyme (``\emph{Alcohol Dehydrogenase}") or a disease (``\emph{Atypical Ductal Hyperplasia}"); ``\emph{Yuri Dolgoruky}" may refer to a Russian prince or a submarine.  In addition, novel or uncommon entities may not exist in the word vocabulary or their embeddings may not be adequately trainable. In order to solve these problems, 
based on Heuristic 2, we propose incorporating linguistic structures to capture context-specific representations of entity mentions.
  


Linguistic structures have been proven useful for capturing the semantics of basic language units~\cite{socher2013recursive,hermann2013role,Socher2013,BowmanPM14}. Considering E2 again, the type of ``\emph{Yuri Dolgoruky}'' can be inferred from its context-specific relational concepts such as ``\emph{nuclear submarines}'', and ``\emph{equip}''. Many linguistic knowledge representations and resources, including AMR (Abstract Meaning Representation~\cite{Banarescu2013}), dependency parsing, semantic role labeling (SRL), VerbNet~\cite{Kipper2006} and FrameNet~\cite{Baker2003}, can be exploited to capture linguistic properties. In this work we focus on AMR to introduce the general framework for specific-context representation generation and later we will compare the impact of various linguistic structures.

AMR is a sembanking language that captures a whole sentence's meaning in a rooted, directed, labeled and (predominantly) acyclic graph structure. AMR represents the semantic structure of a sentence via multi-layer linguistic analyses such as PropBank frames, non-core semantic roles, coreference, named entity annotation, modality, and negation. 
Compared with dependency parsing and SRL, the nodes in AMR are concepts instead of words, and the edge types are much more fine-grained. The AMR language contains rich relations, including frame arguments (e.g., :ARG0, :ARG1), general semantic relations (e.g., :mod, :topic, :domain), relations for quantities, date-entities or lists (e.g., :quant, :date, :op1), and the inverse of all these relations (e.g., :ARG0-of, :quant-of). We carefully select 8 entity-related relation types (:ARG0, :ARG1, :ARG2, :ARG3, :conj, :domain, :topic, :location) from AMR for entity typing. 

\begin{figure}[!htp]
\centering
\includegraphics[width=.48\textwidth]{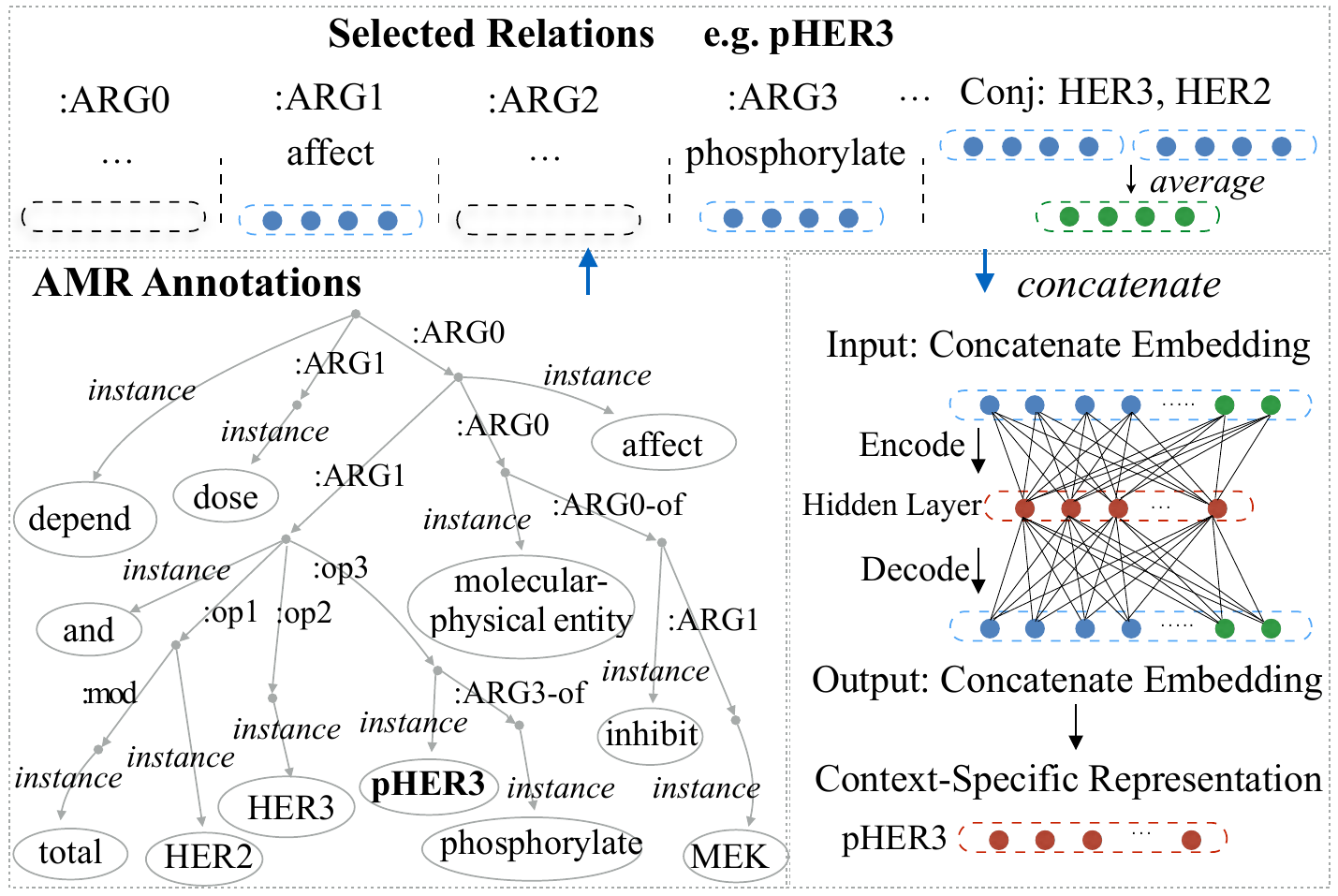}
\caption{Context-Specific Representation Generation for ``pHER3'' based on AMR annotation. (The representations of :ARG1 and :ARG3 are generated based on the general embeddings of ``affect'' and ``phosphorylate'' while the representation of Conj relation is generated based on the average embeddings of ``HER2'' and ``HER3''.)}
\label{specificContext}
\end{figure}

Given an entity mention, we first select its related concepts. For each AMR relation we generate a representation based on the general embeddings of these related concepts. 
If a related concept doesn't exist in the vocabulary, we generate its embedding based on the additive model. If there are several argument concepts involved in a specific relation, we average their representations. We concatenate the vector representations of all selected relations into one single vector. 
Though we have carefully aggregated and selected the popular relation types, the representation of each entity mention is still sparse. To reduce the dimensions and generate a high quality embedding for the specific context, we utilize the sparse auto-encoder framework~\cite{ng2011sparse} to learn more low-dimensional representations. Figure~\ref{specificContext} depicts the context-specific representation generation for ``\emph{pHER3}" in the example E4.

\vspace{-0.2cm}
\vspace{-0.1cm}
\subsection{Knowledge Representation}
\label{sec:kr}


The types of some entities heavily rely on  domain-specific knowledge.  Existing broad-coverage knowledge bases such as DBpedia, Freebase, or YAGO, as well as domain-specific ontologies like BioPortal and NCBO can provide useful knowledge for inferring specific fine-grained types. For example, in DBPedia, both of the properties (e.g., ``\emph{birthPlace}'', ``\emph{party}'' for ``\emph{Mitt Romney}'', ``\emph{capital}'', ``\emph{longitude}'' for ``\emph{Michigan}'') and type labels (e.g., ``\emph{Person}'', ``\emph{Governor}'' for ``\emph{Mitt Romney}'', ``\emph{Place}'', ``\emph{Location}'' for ``\emph{Michigan}'') can be used for entity typing. For the biomedical domain we can consult BioPortal for domain-specific properties and type labels (e.g., ``\emph{Oncogenes}'', ``\emph{Genes}'' for ``\emph{HER2}''). These properties can be used to measure the similarity between mentions. In this work, we construct a knowledge graph based on these properties and type labels and generate knowledge representations for all entities based on a graph embedding framework.

\begin{figure}[!htp]
\centering
\includegraphics[width=.45\textwidth]{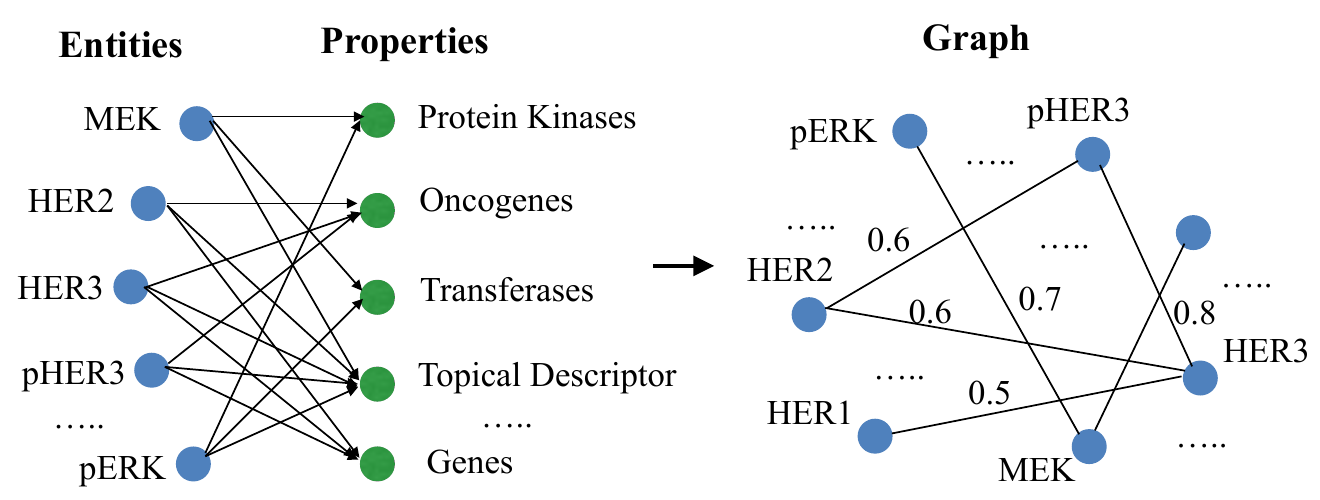}
\caption{Knowledge Graph Construction}
\label{graph}
\vspace{-0.1cm}
\end{figure}

Given an entity $e$, we first collect all properties and type labels $p(e)$ from the domain-specific knowledge bases (KBs). We remove the universal labels from all entities and construct a weighted undirected graph $G=(V, E)$, where $V$ is the set of entities in all KBs and $E$ is the set of edges. As shown in Figure~\ref{graph}, the existence of an edge $e_{ij} = (e_i, e_j)$ implies that two entities, $e_i$ and $e_j$, share some common properties or type labels. The weight of the edge, $w_(e_i, e_j)$, is calculated as:

\vspace{-0.3cm}
\begin{displaymath}
w(e_1, e_2) = \frac{|p(e_1) \cap p(e_2)|}{\max (|p(e_1)|, |p(e_2)|)} 
\end{displaymath}



After constructing the knowledge graph, we apply the graph embedding framework  proposed by~\cite{tang2015line} to generate a knowledge representation for each entity. The basic idea is to optimize the preservation of both local structures, which 
are represented by the observed edges in the graph, and global graph structures, which 
are determined through the shared neighborhood structures of the entities. The global structures assume that entities sharing many neighbors are similar to each other. In this case, each entity is also treated as a specific ``context'' and entities with similar distributions over the ``contexts'' are assumed to be similar. Therefore, each entity plays two roles: the entity itself and a specific ``context'' of other entities. Two objectives that preserve the local and global structures are as follows:

\vspace{-0.2cm}
\begin{displaymath}
O_{local} = -\sum_{(i, j)\in E} w_{ij}log p_{1}(e_i, e_j)
\end{displaymath}
\vspace{-0.2cm}
\begin{displaymath}
p_{1}(e_i, e_j) = \frac{1}{1+exp(-u_{i}^{T}u_{j})}
\end{displaymath}
\vspace{-0.2cm}
\begin{displaymath}
O_{global} = -\sum_{(i, j)\in E} w_{ij}log p_{2}(e_j | e_i)
\end{displaymath}
\vspace{-0.2cm}
\begin{displaymath}
p_{2}(e_j | e_i) = \frac{exp(u_{j}^{'T}u_{i})}{\sum_{k=1}^{|V|}exp(u_{k}^{'T}u_{i})}
\end{displaymath}

where $p_{1}(e_i, e_j)$ defines the joint probability between $e_i$ and $e_j$ and $p_{2}(e_j | e_i)$ defines the probability of ``context'' $e_j$ generated by $e_i$.  $w_{ij}$ is the weight of edge $(e_i, e_j)$. $u_i$ is the low-dimensional vector representation of $e_i$ and $u_{i}^{'}$ is the representation of $e_i$ when it is treated as a specific ``context''. $\{u_{i}\}_{i=1, ..., |V|}$ and $\{u_{i}^{'}\}_{i=1, ..., |V|}$ are all randomly initialized. By tuning $\{u_{i}\}_{i=1, ..., |V|}$ and minimizing the objective function $O_{local}$, we can generate a d-dimensional representation for each entity. To train the function $O_{global}$ efficiently, we adopt a negative sampling approach, which samples multiple negative edges according to some noisy distribution for each edge and specifies the following objective function for each edge $(e_i, e_j)$:

\vspace{-0.4cm}
\begin{displaymath}
log\sigma(u_{j}^{'T}\cdot u_{i}) + \sum_{i=1}^{K}E_{e_{n}\sim P_{n}(e)}[log\sigma(u_{n}^{'T}\cdot u_{i})]
\end{displaymath}

where $\sigma (x) = 1/(1+exp(-x))$ is the sigmoid function. The first term models the observed edges and the second term models the negative edges drawn from the noise distribution and $K$ is the number of negative edges. $P_{n}(e)\propto d_{e}^{3/4}$, where $d_e$ is the out-degree of $e$. To optimize the negative sampling objective function, we adopt the asynchronous stochastic gradient algorithm~\cite{recht2011hogwild}. At each step, the algorithm samples a mini-batch of edges and updates the model parameters. If an edge $(e_i, e_j)$ is sampled, the gradient of the embedding vector $u_{i}$ can be calculated as:

\vspace{-0.3cm}
\begin{displaymath}
\frac{\delta(O_{global})}{\delta(u_{i})} = w_{e_i, e_j}\cdot\frac{\delta log p_{2}(e_j|e_i)}{\delta (u_i)}
\end{displaymath}

After training these two objective functions $O_{local}$ and $O_{global}$ separately, we concatenate the optimized embeddings for each entity as their knowledge representations. 

Next we utilize a domain- and language-independent entity linking system~\cite{Wang2015} to link each mention to existing KBs to determine its knowledge representation. This system is based on an unsupervised collective inference approach. 

Given a set of entity mentions $M = \{m_1, m_2, ..., m_n\}$, this system first constructs a graph for all entity mentions based on their co-occurrence within a paragraph, which, in our experiment, is taken to mean `five continuous sentences'. Each node in the graph denotes an entity mention and two nodes will share an edge if two mentions co-occur within a paragraph. Then, for each entity mention $m$, this system uses the surface form dictionary $<f, {e_1, e_2, ..., e_k}>$, where ${e_1, e_2, ..., e_k}$ is the set of entities with surface form $f$ according to their properties (e.g., labels, names, aliases), to locate a list of candidate entities $e \in E$ and compute the importance score by an entropy based approach~\cite{Zheng2014}. Finally, it computes similarity scores for each entity mention and candidate entity pair $<m, e>$ and selects the most confident candidate (confidence score $>0.95$) as the appropriate entity for linking. If a mention cannot be linked to a KB, we will assign a random vector as its knowledge representation. In our experiments, about $77.7\%$ entity mentions in the general news domain and about $71.4\%$ percent in the biomedical domain can be linked to KBs with high confidence.


\section{Joint Linking, Hierarchical Typing, and Naming}


\subsection{Hierarchical Typing}

\begin{algorithm}[t]
\caption{\small Hierarchical Clustering Algorithm}
\textbf{Input:} \small{mention set $\mathcal{M}$, vector set $\mathcal{V}$, linkable mentions $\Gamma \subseteq \mathcal{M}$, and their type paths $\Delta$} \\
\textbf{Output:} \small{The Hierarchical Clustering Results}
\begin{enumerate}
\item Set the minimum ($K_{min}$) and maximum ($K_{max}$) number of clusters
\item \small{For Layer $i=0$, assuming the initial partition set for $\mathcal{M}$ is $\mathbb{R}^{0}(\mathcal{M}) = \{M\}$}
\item For Layer $i+1$
  \begin{itemize}
    \item $\mathbb{R}^{i+1}(\mathcal{M}) = \mbox{\textbf{Algorithm~\ref{alg}}}(\mathbb{R}^{i}(\mathcal{M}), \mathcal{V}, \Gamma, \Delta, K_{min}, K_{max})$
  \end{itemize}
\item $i = i+1$, repeat step 3, until each cluster $C \in \mathbb{R}^{i+1}(\mathcal{M})$ contains no linkable entity mentions or contains only one mention.
\item return $\{\mathbb{R}^{1}(\mathcal{M}), \mathbb{R}^{2}(\mathcal{M}), \ldots, \mathbb{R}^{N}(\mathcal{M})\}$
\end{enumerate}
\label{alg2}
\end{algorithm}

\begin{algorithm}
\caption{\small Optimal Partition Search}
\textbf{Input:} \small{initial partition set $\mathbb{R}^{i}(\mathcal{M})=\{C_1, C_2, \ldots, C_k\}$, vector set $\mathcal{V}$, linkable mentions and their type paths $\Gamma$, $\Delta$, the minimal ($K_{min}$) and maximal ($K_{max}$) number of clusters for partition} \\
\textbf{Output:} \small{the optimal parameters: $w_1^*, w_2^*, w_3^*$; and the optimal next layer partition set $\mathbb{R}^{i+1}(\mathcal{M})$}
\begin{itemize}
\item $o_{min} = \infty$; $w_1^*=w_2^*=w_3^*=0.33$; $\mathbb{R}^{i+1}(\mathcal{M})$;
\item For $w_1 = 0$ to $w_1 = 1$ by steps of 0.05
  \begin{itemize}
  \item For $w_2 = 0$ to $w_2 = 1$ by steps of 0.05 such that $w_1+w_2 \leq 1.0$
    \begin{itemize}
    \item $w_3 = 1.0-w_1-w_2$
	\item $\mathbb{R}_{curr}(\mathcal{M}) = \bigcup_{C_i \in \mathbb{R}^{i}(\mathcal{M})} Xmeans_{w_1,w_2,w_3}(C_i)$
    \item $o_{curr} = O(\mathbb{R}_{curr}(\mathcal{M}), \Gamma, \Delta)$
    \item if $o_{curr} < o_{min}$:
      \begin{itemize}
      \item $o_{min} = o_{curr}$; $w_1^*=w_1$; $w_2^*=w_2$; $w_3^*=w_3$
      \item $\mathbb{R}^{i+1}(\mathcal{M}) = \mathbb{R}_{curr}(\mathcal{M})$
      \end{itemize}
    \end{itemize}
  \end{itemize}
\item return $w_1^*, w_2^*, w_3^*$; $\mathbb{R}^{i+1}(\mathcal{M})$;
\end{itemize}

\label{alg}
\end{algorithm}

For an entity mention $m \in \mathcal{M}$, the vector representation $v$  of $m$ is the concatenation of the three parts mentioned above: the distributed general semantic representation $v_{E}$, the local-context specific representation $v_{C}$ and the knowledge representation $v_{K}$ based on entity linking. We designed a hierarchical X-means clustering algorithm to detect the hierarchical types of entities. X-means~\cite{Pelleg2000} is an extension of the well-known K-means algorithm for which the number of clusters is estimated instead of being fixed by the user. It has two major enhancements compared to K-means: 1) It is fast and scales great with respect to the time it takes to complete each iteration; 2) It can automatically estimate the number of clusters and also obtain local and global optimals for specific data sets.

X-means requires a distance metric $D$ to calculate its clusters. We define $D$ given two entity mentions $m_1$ and $m_2$ with vector representations $v_1$ and $v_2$ respectively (we regard the vector of the cluster centroid as the same as the vector of the mention), as:

\vspace{-0.3cm}
\begin{equation*}
\begin{split}
D(m_1, m_2) = w_1\centerdot D(v_{E}^{m_1}, v_{E}^{m_2}) \\
+ w_2\centerdot D(v_{C}^{m_1}, v_{C}^{m_2})+ w_3\centerdot D(v_{K}^{m_1}, v_{K}^{m_2})
\end{split}
\end{equation*}

Here, $D(\centerdot)$ is the Euclidean distance between two vectors. $w_1$, $w_2$, $w_3$ are the balance parameters among three types of representations and $w_1 + w_2 + w_3 = 1$. 

Given the set of all mentions $\mathcal{M}$, we will select highly linkable mentions (confidence score $>0.95$) $\Gamma \subseteq \mathcal{M}$ and their corresponding type paths $\Delta$ based on the entity linking system described in Section~\ref{sec:kr} for typing and naming. Here, the type path denotes the longest path from the KB title to the root of the type hierarchy in the KB. For example, we can link the entity mention ``\textbf{Mitt Romney}'' in Section~\ref{sec:introduction} to YAGO and extract a type path from the entity title to the root: `\emph{Mitt Romney} $\rightarrow$ \emph{Governor} $\rightarrow$ \emph{Politician}$ \rightarrow $\emph{Leader}$ \rightarrow $\emph{Person}  $\rightarrow$ \emph{Entity}.' As outlined in Algorithm~\ref{alg2}, we start from the initial set of all entity mentions $\mathcal{M}$ and vector representations $\mathcal{V}$ to generate hierarchical partitions $\{\mathbb{R}^{1}(\mathcal{M}), \mathbb{R}^{2}(\mathcal{M}), \ldots, \mathbb{R}^{N}(\mathcal{M})\}$, where $\mathbb{R}^{i}(\mathcal{M})$ represents the partition of $\mathcal{M}$ based on vector representation set $\mathcal{V}$ at layer $i$.





For each layer $i$, in order to get further partition set $\mathbb{R}^{i+1}(\mathcal{M})$ based on $\mathbb{R}^{i}(\mathcal{M})= \{C_1, C_2, \ldots, C_k\}$, we define $Xmeans_{w_1,w_2,w_3}(C)$ as the partition of mention set $C$ based on running X-means with $D$ parameterized by the parameter set $w_1, w_2, w_3$. It remains to search for the optimal $w_1, w_2, w_3$. In order to judge an optimal partition for each layer, we utilize information from the KB: $\Gamma$ and $\Delta$, as truth and invoke the following heuristic:

\vspace{+0.2cm}
\textbf{Heuristic 4: }\textbf{\emph{The optimal clustering results of all mentions should be consistent with the optimal clustering results of all linkable mentions.}}
\vspace{+0.2cm}


 We then define an objective function $O$ that evaluates a certain layer of partition set $\mathbb{R}(\mathcal{M}) = \{C_1, C_2, \ldots, C_k\}$:

\begin{displaymath}
O (\mathbb{R}, \Gamma, \Delta)= \mathbb{D}_{inter} + \mathbb{D}_{intra}
\end{displaymath}
\vspace{-0.3cm}
\begin{displaymath}
\mathbb{D}_{inter} = \sum_{i\neq j =1}^{n}\sum_{m_{u}^{'}\in C_{i}, m_{v}^{'}\in C_{j}} w(m_{u}^{'}, m_{v}^{'})
\end{displaymath}
\vspace{-0.3cm}
\begin{displaymath}
\mathbb{D}_{intra} =  \sum_{i=1}^{n}\sum_{m_{u}^{'},m_{v}^{'}\in C_{i}}(1-w(m_{u}^{'}, m_{v}^{'}))
\end{displaymath}

\begin{table*}[!htp]
\centering
\begin{tabular}{p{2.0cm}|p{1.5cm}|p{1.5cm}|p{1.5cm}|p{1.5cm}|p{1.3cm}|p{1.2cm}|p{1.5cm}}
\hline
 & English News & Biomedical Articles & Discussion Forum & Chinese News & Japanese News & Hausa News & Yoruba News\\
 \hline
 \# of docs & 367 & 14 & 329 & 20 & 25 & 90 & 239\\
 \hline
 \# of mentions & 15002 & 2055 & 4157 & 1683 & 489 & 1508 & 7456 \\
 \hline
 \# of types & 183 & 51 & 149 & 44 & 47 & 3 & 3\\
 \hline
\end{tabular}
\caption{Statistics of Test Data sets}
\label{dataset}
\vspace{-0.1cm}
\end{table*}

where $w(\centerdot)$ is defined in Section~\ref{sec:kr} based on type paths. Algorithm~\ref{alg} encapsulates the search for $w_1^*, w_2^*, w_3^*$, the parameter set that optimizes $O$.

\subsection{Hierarchical Type Naming}
\label{sec:typeNaming}

The entity linking system described in Section~\ref{sec:kr} can extract highly linkable entity mentions and their corresponding type name paths. Considering the examples in Section~\ref{sec:introduction} again, we can link the entity mention ``\textbf{Mitt Romney}'' to YAGO and extract a type path from the entity title to the root: `\emph{Mitt Romney} $\rightarrow$ \emph{Governor} $\rightarrow$ \emph{Politician}$ \rightarrow $\emph{Leader}$ \rightarrow $\emph{Person}  $\rightarrow$ \emph{Entity}.' Similarly, we can link ``\textbf{HER2}'' to ``\emph{ERBB2}'' in BioPortal and extract the type name path from the entity to the root of an ontology as: `\emph{ERBB2} $\rightarrow$ \emph{Proto-Oncogenes} $\rightarrow$ \emph{Oncogenes} $\rightarrow$ \emph{Genes} $\rightarrow$ \emph{Genome Components} $\rightarrow$ \emph{Genome} $\rightarrow$ \emph{Phenomena and Processes} $\rightarrow$ \emph{Topical Descriptor} $\rightarrow$ \emph{MeSH Descriptors}'. We first normalize the type name paths and remove those too general type name candidates (e.g., `\emph{Entity}', `\emph{Topical Descriptor}'). In our experiments, a type name is removed if more than 90\% of type paths contain it. Then, we generate the most confident type label $n_C$ for each cluster $C$ based on high confidence linking results as follows:  

For a specific cluster $C$, the mentions within this cluster are denoted as $M_{C}$ and the highly linkable mentions are $\Gamma_{C} \subseteq M_C$. We collect all the type names $N_{C}$ from the type paths of all $m \in \Gamma_{C}$, then we determine which type name is the most fine-grained and also matches with cluster $C$ based on two metrics:  \emph{Majority} and \emph{Specificity}. Majority is measured based on the frequency of the specific type name $n$ in the type name set $N_{C}$. This metric is designed based on our intuition that the type name should be able to represent the types of as many entity mentions as possible. Specificity is designed to measure the granularity degree of the type name in the whole name path. These two metrics are computed as follows:

\begin{displaymath}
majority_{n}^{C} = \frac{Count (n, C)}{|M_{C}|}
\end{displaymath}
\begin{displaymath}
specificity_{n}^{p(n)} = \frac{Position(n, p(n))}{|p(n)|}
\end{displaymath}

\noindent where $Count (n, C)$ represents the frequency of a type name $n$ in the set $N_{C}$, $|M_{C}|$ represents the number of members in cluster $C$, $p(n)$ represents the longest type name path including $n$ and $Position(n, p(n))$ represents the position of $n$ in $p(n)$ (from the root to $n$). 

We combine these two metrics and choose $n_C$ as follows: 
for each cluster $C$ we define $N_{C}^{m} = \{n : n \in N_{C} \wedge majority_n^{C} \ge \lambda \}$,
where $\lambda$ is a threshold parameter\footnote{We set $\lambda$ to 0.75 in our experiments.}. 
We then select $n_C = \displaystyle\argmax_{n \in N_C^m} specificity_n^{p(n)}$.
For example, if the majority of `\emph{Proto-Oncogenes}' and `\emph{Genes}' are both larger than $\lambda$, which is set as 0.75 in our experiments, we should choose `\emph{Proto-Oncogenes}' because it is much more fine-grained than `\emph{Genes}' in the whole type name path. After naming for each hierarchical cluster, we will generate a type hierarchy, which is also customized for the specific corpus.

\section{Experiments and Evaluation}





In this section we present an evaluation of the proposed framework on various genres, domains and languages, as well as a comparison with state-of-the-art systems.

\subsection{Data}

We first introduce the data sets for our experiments. To compare the performance of our framework against state-of-the-art name taggers and evaluate its effectiveness on various domains and genres, we first conduct experiments on Abstract Meaning Representation (AMR)  data sets, which include perfect mention boundaries with  fine-grained entity types. For the experiment on multiple languages, we use data sets from the DARPA LORELEI program\footnote{http://www.darpa.mil/program/low-resource-languages-for-emergent-incidents} and foreign news agencies. The detailed data statistics are summarized in Table~\ref{dataset}.


Since our approach is based on word embeddings, which need to be trained from a large corpus of unlabeled in-domain articles, we collect all the English and Japanese articles from the August 11, 2014 Wikipedia dump to learn English and Japanese word/phrase embeddings and collect all the articles of the 4th edition of the Chinese Gigaword Corpus\footnote{https://catalog.ldc.upenn.edu/LDC2009T27} to learn Chinese word/phrase embeddings. For the biomedical domain, we utilize the word2vec model\footnote{http://bio.nlplab.org/} which is trained based on all paper abstracts from PubMed\footnote{http://www.ncbi.nlm.nih.gov/pubmed} and full-text documents from the PubMed Central Open Access subset. We also collect all entities and their properties and type labels from DBpedia and 300+ biomedical domain specific ontologies crawled from BioPortal~\cite{Zheng2014} to learn knowledge embeddings.
\subsection{Evaluation Metrics}


\label{secmetric}
Our framework can automatically discover new fine-grained types. Therefore, in addition to mention-level Precision, Recall and F-measure, we also exploit the following standard clustering measures:

1) Purity: To compute purity, each system cluster is assigned to the reference class with the greatest mention overlap. The sum off all matching mentions given this assignment is then divided by N.

\vspace{-0.2cm}
\begin{displaymath}
purity = \frac{\sum_{j=1}^{K}\max_{1\leq i\leq M}|C_j \cap R_{i}|}{N} 
\end{displaymath}

where N is the total number of mentions, K is the number of system generated clusters, M is the number of clusters in the answer key, $C_j$ is the set of mentions in the $j^{th}$ cluster in our system, and $R_i$ is the set of mentions for the $i^{th}$ type in the answer key.

2) Precision, Recall, F-measure (F): 
Here, we utilize F-measure to evaluate the entity mentions that match with the ground truth dataset. 

\vspace{-0.3cm}
\begin{displaymath}
prec(R_i, C_j) = \frac{|R_i \cap C_j|}{|C_j|} 
\end{displaymath}
\vspace{-0.3cm}
\begin{displaymath}
rec(R_i, C_j) = \frac{|R_i \cap C_j|}{|R_i|}
\end{displaymath}
\vspace{-0.3cm}
\begin{displaymath}
F_{measure}(R_i, C_j) = \frac{2\ast prec(R_i, C_j)\ast rec(R_i, C_j)}{prec(R_i, C_j) + rec(R_i, C_j)} 
\end{displaymath}

Intuitively, F-measure{\small ($R_i, C_j$)} measures how good a class {\small $R_i$} can be described by a cluster {\small $C_j$} and the success of capturing a class {\small $R_i$} is measured by using the ``best'' cluster {\small $C_j$} for {\small $R_i$}, i.e., the {\small $C_j$} that maximizes {\small $F_{measure}(R_i, C_j)$}.

\begin{displaymath}
F_{measure} = \frac{\sum_{i=1}^{M}|R_i|\max_{1\leq j\leq K}F_{measure}(R_i, C_j)}{N} 
\end{displaymath}
3) Entropy:
The entropy measures the diversity of a cluster {\small $C_j$} and it is defined as

\vspace{-0.5cm}
\begin{displaymath}
entropy(C_j) = -\sum_{i=1}^{M}P(i,j)\ast logP(i,j)
\end{displaymath}
\vspace{-0.2cm}
\begin{displaymath}
P(i, j) = \frac{|R_i \cap C_j|}{N}
\end{displaymath}

where {\small $P(i,j) 
$} represents the probability of a mention in the key cluster {\small $R_i$} being assigned to the system cluster {\small $C_j$}. The lower the value of entropy, the better the clustering is. The overall cluster entropy is then computed by averaging the entropy of all clusters:

\vspace{-0.2cm}
\begin{displaymath}
entropy = \frac{\sum_{j=1}^{K}|C_j|\ast entropy(C_j)}{N}
\end{displaymath}
\subsection{Comparison with State-of-the-art}
\label{sec:baseline}
We compare with two high-performing name taggers, Stanford NER~\cite{finkel2005incorporating} and FIGER~\cite{Ling12fine-grainedentity}, on both coarse-grained types (Person, Location, and Organization), and fine-grained types. We utilize the AMR parser developed by~\cite{flanigan2014discriminative} and manually map AMR types and system generated types to three coarse-grained types.  In order to compare identification results, we design a simple mention boundary detection approach based on capitalization features and part-of-speech features. We compare the performance of our system with both perfect AMR and system AMR annotations with the performance of NER and FIGER. We conduct the experiments on English news data set and link entity mentions to DBPedia~\cite{panunsupervised}. 
The mention-level F-scores are shown in Table~\ref{baseline1}.



\begin{table}[!htb]
\small
\centering
\begin{tabular}{p{1.6cm}|p{1.1cm}|p{1.1cm}|p{1.0cm}|p{1.0cm}}
\hline
Layer (\# of Clusters)& $System^{1}$ & $System^{2}$ & Stanford NER & $FIGER^{1}$ \\
\hline
L1 (5) & 0.649 & 0.628 & \multirow{5}{*}{0.712} & \multirow{5}{*}{0.663}  \\
L2 (21) & 0.668 & 0.647 & & \\
L3 (92) & 0.689 & 0.681 & & \\
L4 (146) & 0.713 & 0.709 & & \\
L5 (201) & 0.728 & 0.721 & & \\
\hline
\end{tabular}
\caption{Coarse-grained Mention-level F-score Comparison ($System^{1}$: based on perfect AMR, $System^{2}$: based on system AMR)}
\label{baseline1}
\end{table}
\vspace{-0.2cm}

\begin{figure}[ht]
\centering
\includegraphics[width=.35\textwidth]{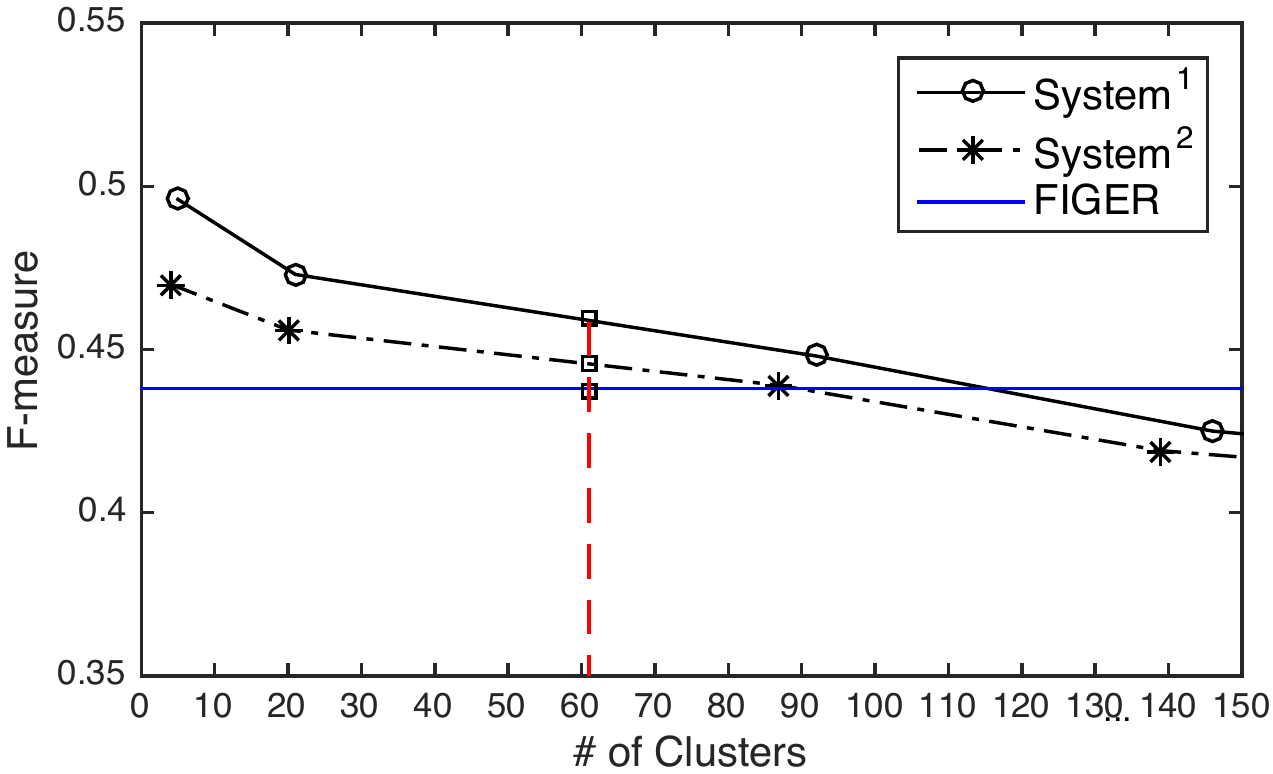}
\caption{Fine-grained Cluster-level F-score Comparison (The red dashed line shows the same \# of clusters for comparison.)}
\label{baseline2}
\vspace{-0.2cm}
\end{figure}

Besides these three coarse-grained types, there are also many new types (e.g., Ailment, Vehicle, Medium) discovered by fine-grained entity typing approaches. We compare our framework with FIGER based on their pre-trained 112-classes classification model. The cluster-level F-scores are shown in Figure~\ref{baseline2}. 

From Table~\ref{baseline1} we can see that, on coarse-grained level, compared with Stanford NER, which contains many features and is trained on about 945 annotated documents (approximately 203,000 tokens), our approach with both system AMR and perfect AMR achieved comparable performance. Compared with FIGER on coarse-grained level, our approach with system AMR and perfect AMR also achieved better results.  
Figure~\ref{baseline2} shows the fine-grained level performance. The number of clusters, to some extent, can reflect the granularity of fine-grained typing. Although we can not directly map the granularity of FIGER to our system, considering the classification results of FIGER are highly biased toward a certain subset of types (about 60 types), our approach with both system AMR and perfect AMR should slightly outperform FIGER, which is trained based on 2 million labeled sentences.

Both Stanford NER and FIGER heavily rely on linguistic features, such as tokens, context n-grams, part-of-speech tags, to predict entity types. Compared with lexical information, semantic information is more indicative to infer its type. For example, in ``\emph{\textbf{Bernama} said Malaysia had procured very short-range air defense systems from Russia.}'' ``\emph{Bernama}'' is assigned the type ``\emph{Person}'' by the FIGER system. However, based on general semantic information, the most similar concepts to ``\emph{Bernama}'' include ``\emph{Malaysiakini,}'' ``\emph{Utusan,}'' and ``\emph{Kompas,}'' which can effectively help infer the correct type as ``\emph{News Agency.}'' In addition, in many cases, the fine-grained types of entity mentions are heavily depend on their knowledge information. For example, in ``\emph{\textbf{Antonis Samaras} is cheered by supporters after his statement in Athens June 17, 2012.}'' it's difficult to infer the fine-grained type of ``\emph{Antonis Samaras}'' based on context words. However, we can utilize more knowledge from KBs and find that the most similar concepts to ``\emph{Antonis Samaras}'' include ``\emph{Kostas Karamanlis,}'' ``\emph{Georgios Papastamkos,}'' and ``\emph{Giannis Valinakis}'' based on knowledge representation, which can help infer the fine-grained type of \emph{Antonis Samaras} as ``\emph{Politician}.'' 

\subsection{Comparison on Genres}
\label{sec:genre}

\begin{figure*}[htp]
\centering
\includegraphics[width=.3\textwidth]{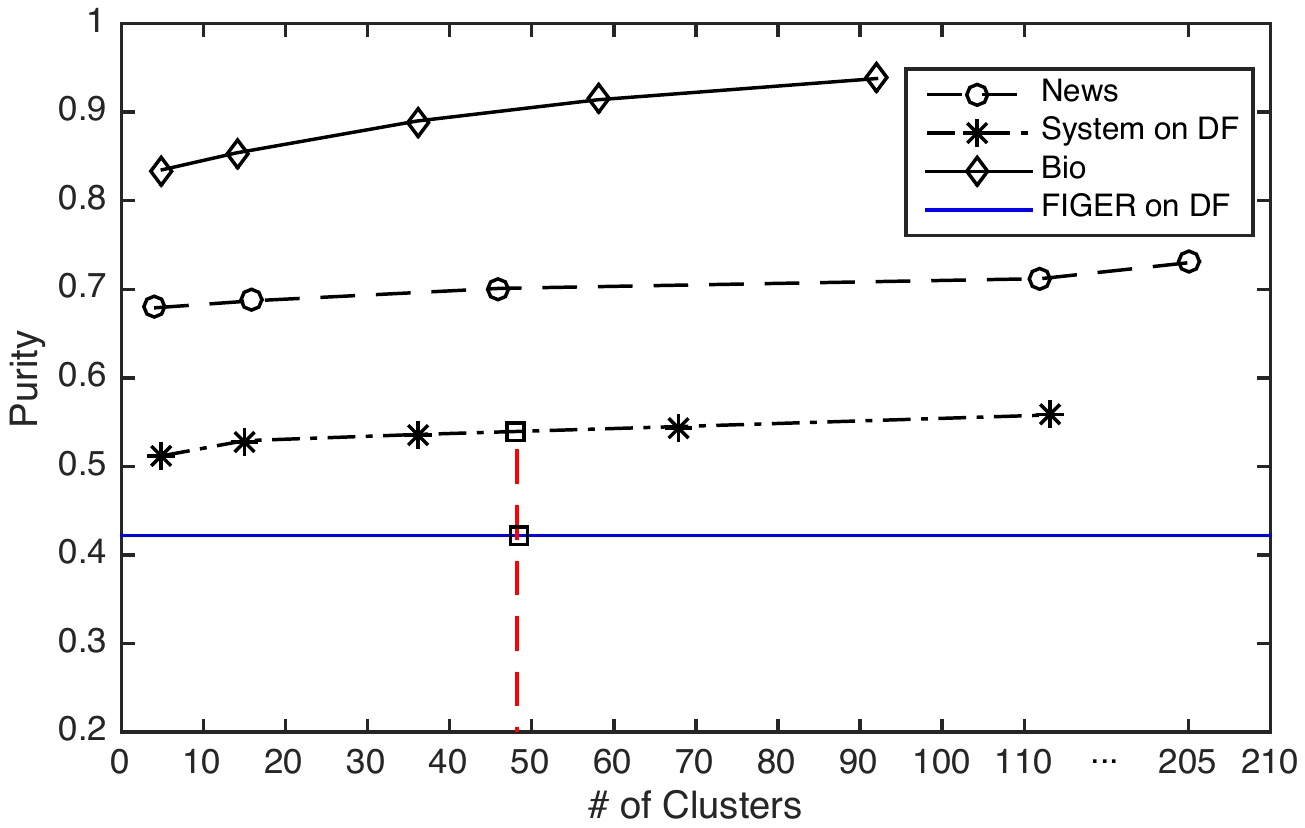}
\includegraphics[width=.3\textwidth]{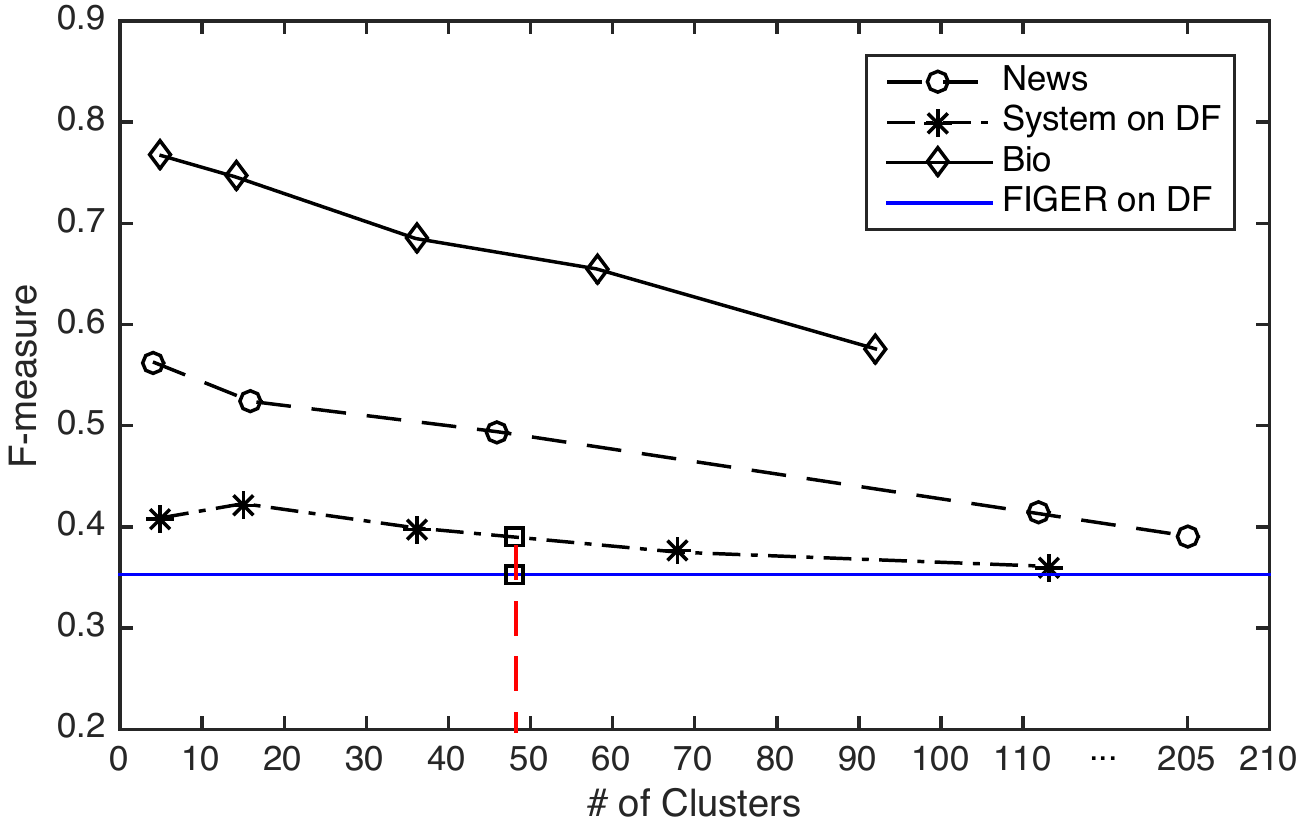}
\includegraphics[width=.3\textwidth]{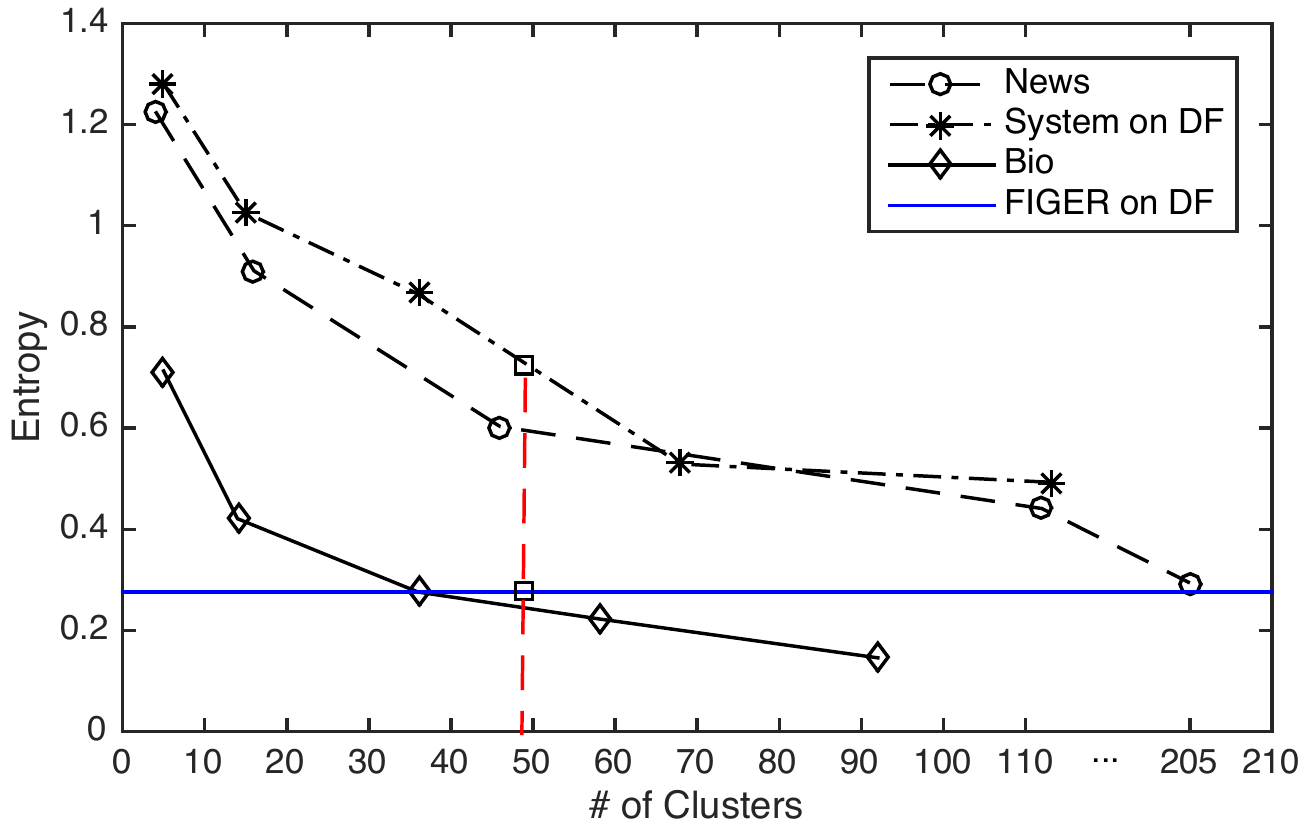}
\caption{Typing Results for Different Genres and Domains (English) with Perfect AMR ((The red dashed line shows the same \# of clusters for comparison.))}
\label{genreResults}
\end{figure*}

For comparison between news and discussion forum genres, we utilize perfect entity boundaries and perfect AMR annotation results to model local contexts and 
link entity mentions to DBpedia~\cite{panunsupervised}. 
Figure~\ref{genreResults} shows the performance. 
We can see that our system performs much better on news articles than discussion forum posts, due to two reasons: (1) many entities occur as abbreviations in discussion forum posts, which brings challenges to both entity typing and linking. For example, in the following post: ``\emph{The joke will be on \textbf{House} \textbf{Dems} who are being promised a bill to ``fi'' the problems with the \textbf{Senate} bill.}'', it's difficult to generate accurate general semantic and knowledge representations for the mentions like ``\emph{House}'' (which refers to ``\emph{United States House of Representatives}'') and ``\emph{Dems}'' (which refers to ``\emph{Democratic Party of United States}''). (2) more novel and uncommon entities appear in discussion forums. Take the following sentence as an example: ``\emph{Mend some fences and get this country moving. He could call it \textbf{APOLOGIES ON BEER}. Hell, sell tickets and hire the Chinese to cater the event.}'' ``\emph{APOLOGIES ON BEER}'' is a novel emerging entity, thus it will be difficult to predict its fine-grained type ``\emph{tour},'' even with semantic and knowledge representations.

In addition, our system can outperform the FIGER system (with gold segmentation), of which the results are focused on about 50 types on the discussion forum data set, on both Purity and F-measure. As discussed in Section~\ref{sec:baseline}, FIGER is trained based on a rich set of linguistic features. When it is applied to a new informal genre, feature generation can not be guaranteed to work well. Our system is mainly based on semantic representations, which will not be affected by the noise.

\subsection{Comparison on Domains}
\label{sec:domain}

To demonstrate the domain portability of our framework, we take the biomedical domain as a case study. For fair comparison, we used perfect AMR semantic graphs and perfect mention boundaries. 
Figure~\ref{genreResults} compares the performance for news and biomedical articles.

%
As shown in Figure~\ref{genreResults}, our system performs much better on biomedical data than on general news data. In an in-depth analysis of the experiment results, we found that most of the entity mentions in the biomedical domain are unique and unambiguous, and the mentions with the same type often share the same name strings. For example, ``\emph{HER2},'' ``\emph{HER3},'' and ``\emph{HER4}'' refer to similar ``\emph{Proto-Oncogenes}''; ``\emph{A-RAF},'' ``\emph{B-RAF}'' and ``\emph{C-RAF}'' share the same type ``\emph{RAF Kinases}.'' However, it is always the opposite in the general news domain. For example, although ``\emph{Shenzhen},'' ``\emph{Shenzhen Maoye},'' ``\emph{Shenzhen Gymnasium}'' share the same name string ``Shenzhen'', they have different entity types: ``\emph{Shenzhen}'' refers to a city, ``\emph{Shenzhen Maoye}'' is a company and ``\emph{Shenzhen Gymnasium}'' is a facility. What's more, ambiguity commonly exists in general news domain, especially for persons and locations. For example, both of``\emph{Sokolov},'' ``\emph{Chamberlain}.'' can refer to a city or a person.  We utilize the ambiguity measure defined in~\cite{ji2010overview} as the criteria to demonstrate the ambiguity degree of news and biomedical domains.

\vspace{-0.2cm}
\begin{displaymath}
ambiguity = \frac{\#name\ strings\ belong\ to\ more\ than\ one\ cluster}{\#name\ strings} 
\end{displaymath}

\begin{figure}[h]
\centering
\includegraphics[width=.37\textwidth]{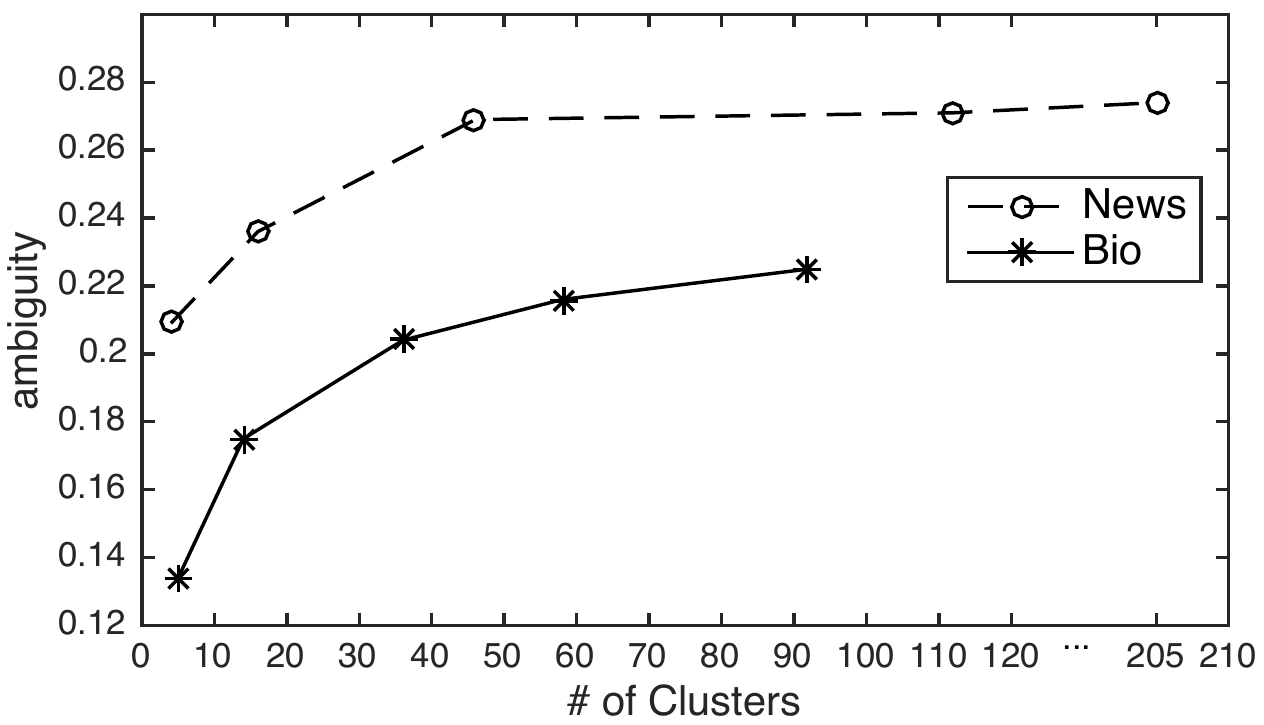}
\caption{Ambiguity Comparison for Different Domains}
\label{ambiguity}
\end{figure}

Figure~\ref{ambiguity} shows the ambiguity comparison results between the general news and biomedical domains. Due to the low ambiguity of the biomedical domain, the general semantic representation and knowledge representation can better capture the domain-specific types of these entity mentions. This analysis can also be verified by the final optimal weights for three kinds of representations: $w_1=0.45, w_2=0.05, w_3=0.5$ for biomedical domain while $w_1=0.45, w_2=0.2, w_3=0.35$ for news, which shows the different contributions of three-layer representations for entity typing.





\subsection{Comparison on Languages}

\begin{figure*}[htp]
\centering

\includegraphics[width=.3\textwidth]{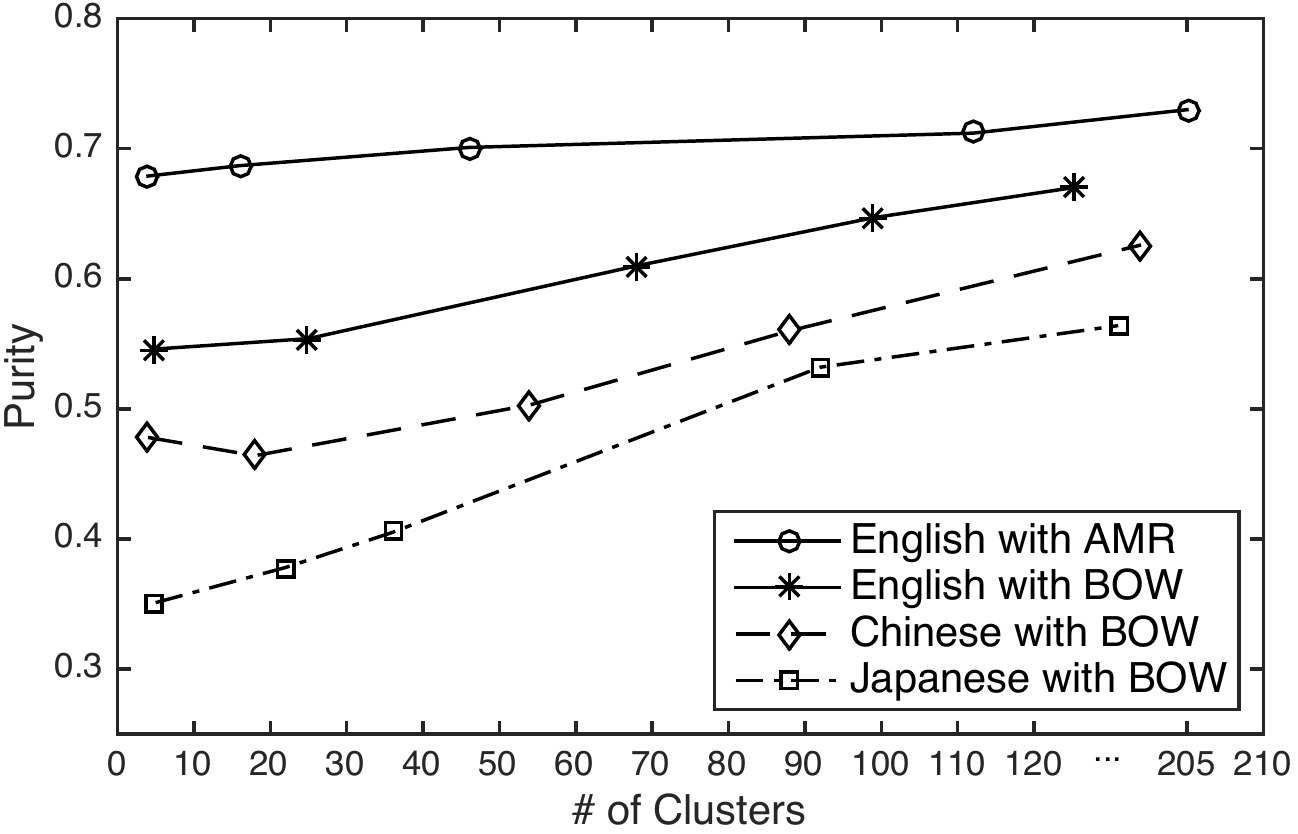}
\includegraphics[width=.3\textwidth]{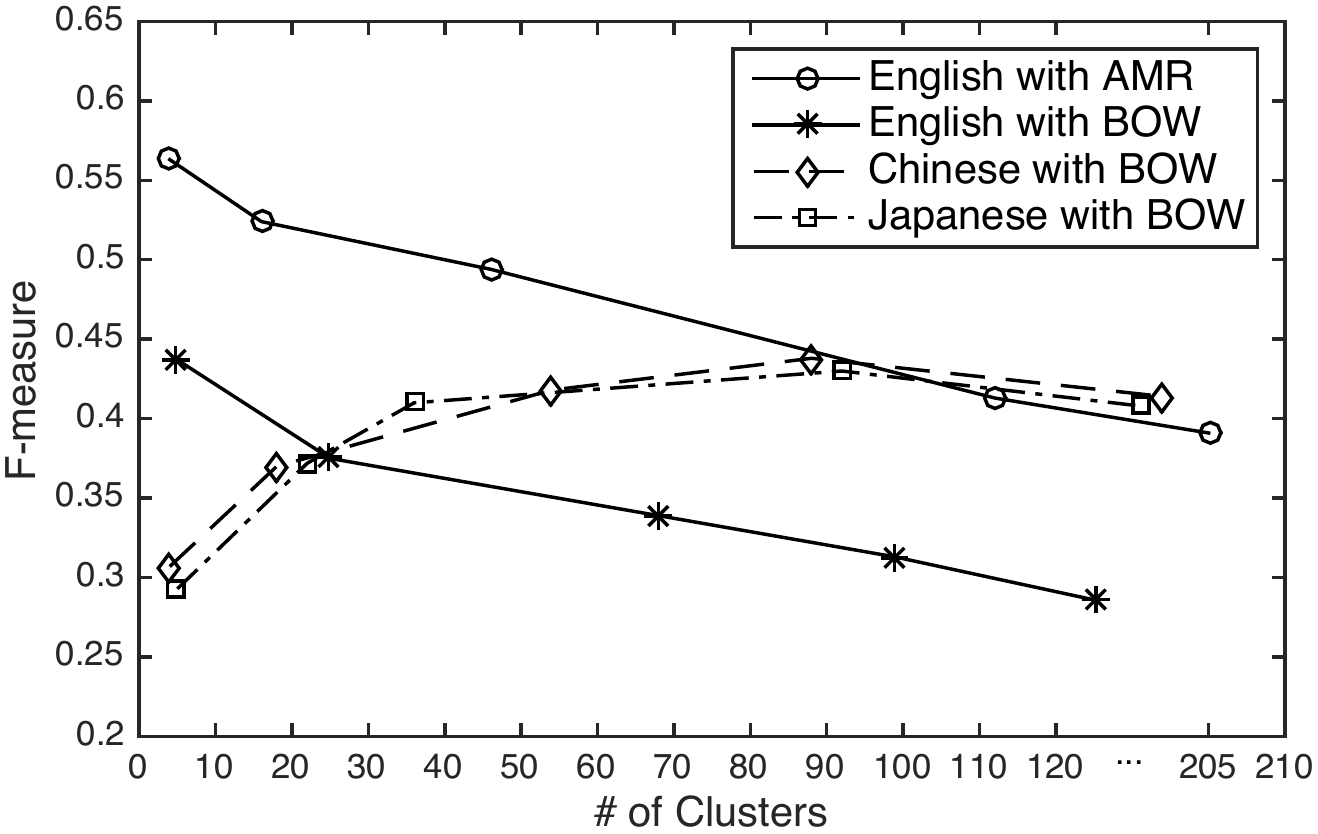}
\includegraphics[width=.3\textwidth]{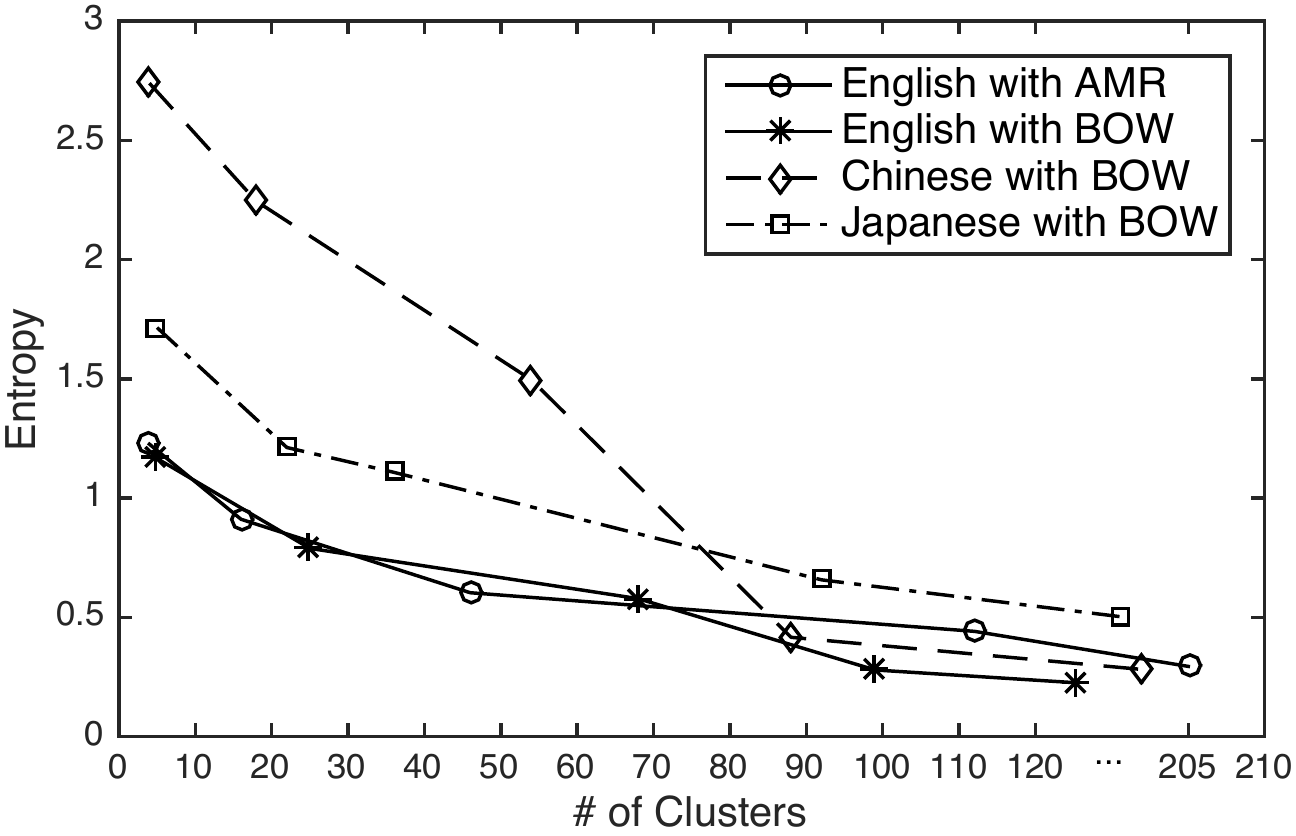}

\caption{Typing Results for Various Languages (News)}
\label{languageResults}
\end{figure*}

\begin{figure*}[htp]
\centering
\includegraphics[width=.3\textwidth]{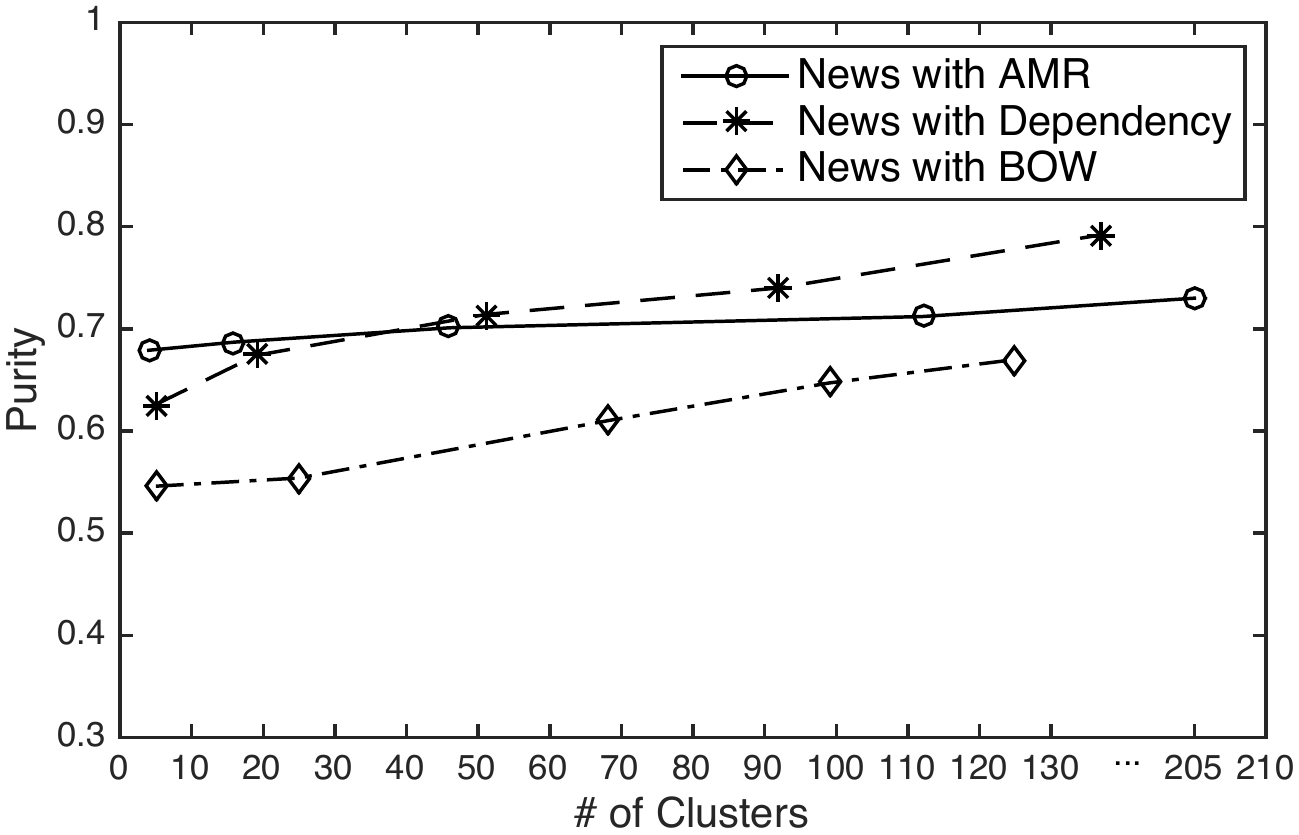}
\includegraphics[width=.30\textwidth]{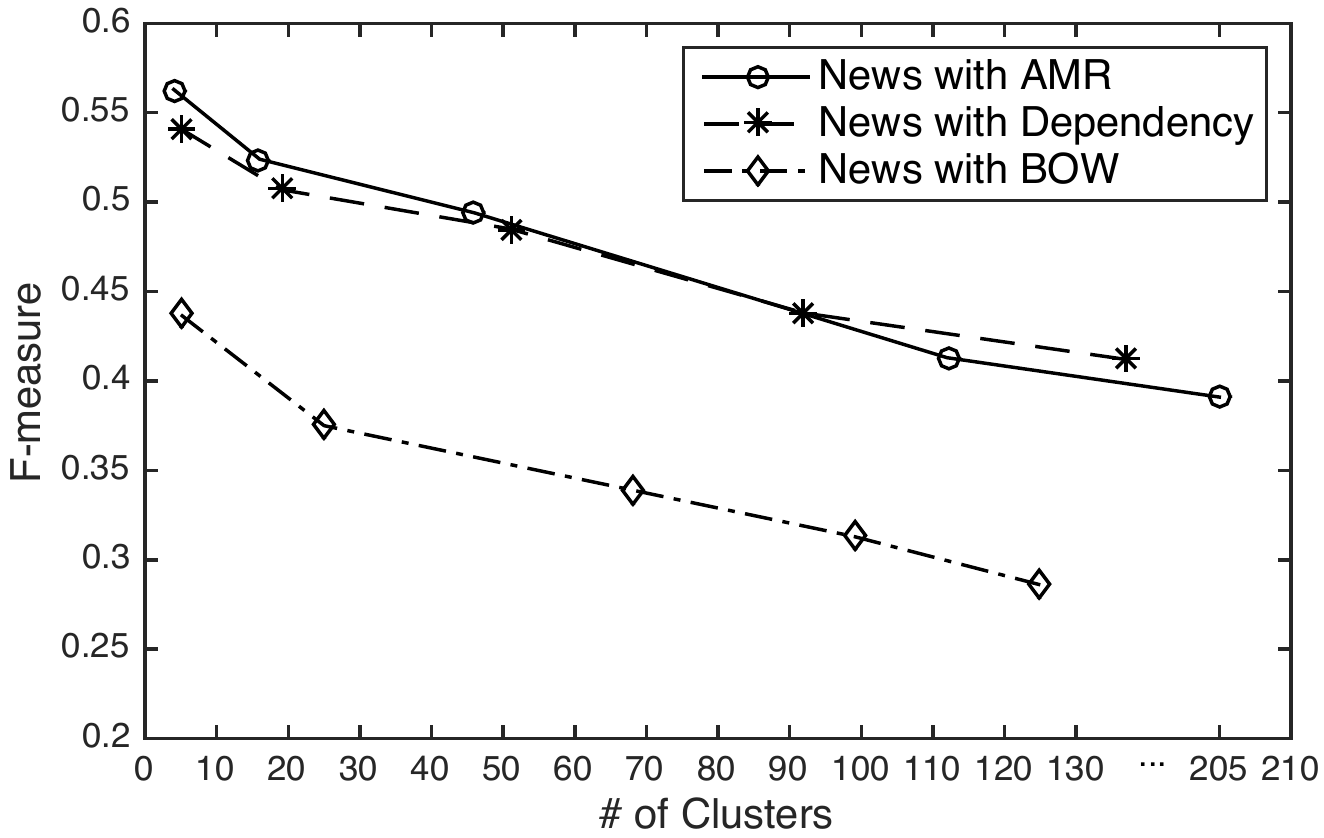}
\includegraphics[width=.3\textwidth]{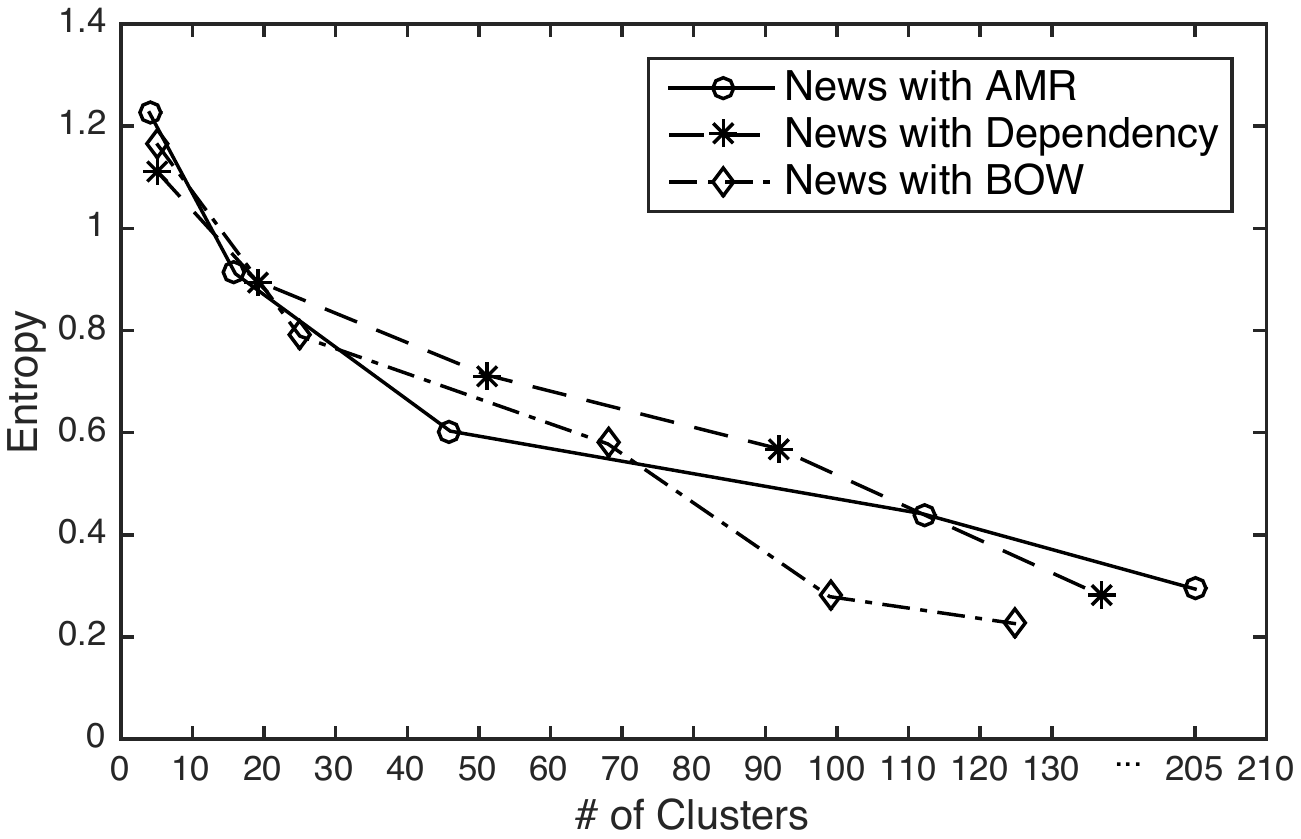}

\caption{Linguistic Structures for Context Specific Representation}
\label{linguistic}
\end{figure*}

Our framework is also highly portable to new languages. Different languages may have different linguistic resources available. For example, English has rich linguistic resources (e.g., AMR) that can be utilized to model local contexts while some languages (e.g., Chinese, Japanese) don't. To evaluate the impact of the local contexts on entity typing, we compare the performance based on AMR and the embeddings of context words that occur within a limited-size window. In our experiment, the window size is 6. Figure~\ref{languageResults} shows the performance on English, Chinese and Japanese news data sets.

Figure~\ref{languageResults} shows that our framework on Chinese and Japanese also achieved comparable performance as English. The main reason is that entities in Chinese and Japanese have less ambiguity than English. Almost all of the same name strings refer to the same type of entity. Based on the ambiguity measure in Section~\ref{sec:domain}, the ambiguity is lower than 0.05 both for Chinese and Japanese.

In addition, for low resource languages, there are not enough unlabeled documents to train word embeddings and KBs may not be available for these languages. In this case, we can utilize other feature representations such as bag-of-words tf-idf instead of embedding based representations. To prove this, we apply our framework to two low-resource languages: Hausa and Yoruba. The mention-level typing accuracy with perfect boundary is very promising: $85.42\%$ for Hausa and $72.26\%$ for Yoruba. 


\subsection{Comparison on Linguistic Structures}
\label{sec:alternatesemantics}

In order to evaluate the impact of context-specific representations on entity typing, we compare the performance based on various linguistic structures, including AMR, dependency relations, and bag-of-words in a specific window. 
Figure~\ref{linguistic} presents the performance on English news dataset.
The general entity semantic representation and knowledge representation are the same for all these three sets of experiments.

Compared with bag-of-words, both AMR and dependency based context-specific representations can achieve better performance, which demonstrates the importance of incorporating a wider range of deep knowledge and semantic relations for context representation. Specifically,  AMR and dependency relations can more effectively capture meaningful context information than bag-of-words. For example, in the sentence ``\emph{The \textbf{Kuwolsan} was rumored to be \underline{carrying} \underline{arms} or \underline{ammunition} possibly intended for Pakistan,}''  \emph{Kuwolsan} is an out-of-vocabulary concept. The bag-of-words based method will generate the context-specific representation based on the words like ``\emph{the, was, rumored, to, be, carrying}'', which is not useful for inferring the type. However, both AMR and dependency based methods can capture semantically related concepts ``\emph{carrying, arms, ammunition}'', which will help the system to infer the type of ``\emph{Kuwolsan}'' as ``\emph{Ship}''.

\subsection{Type Naming Evaluation}

\begin{figure}[htp]
\centering
\includegraphics[width=.37\textwidth]{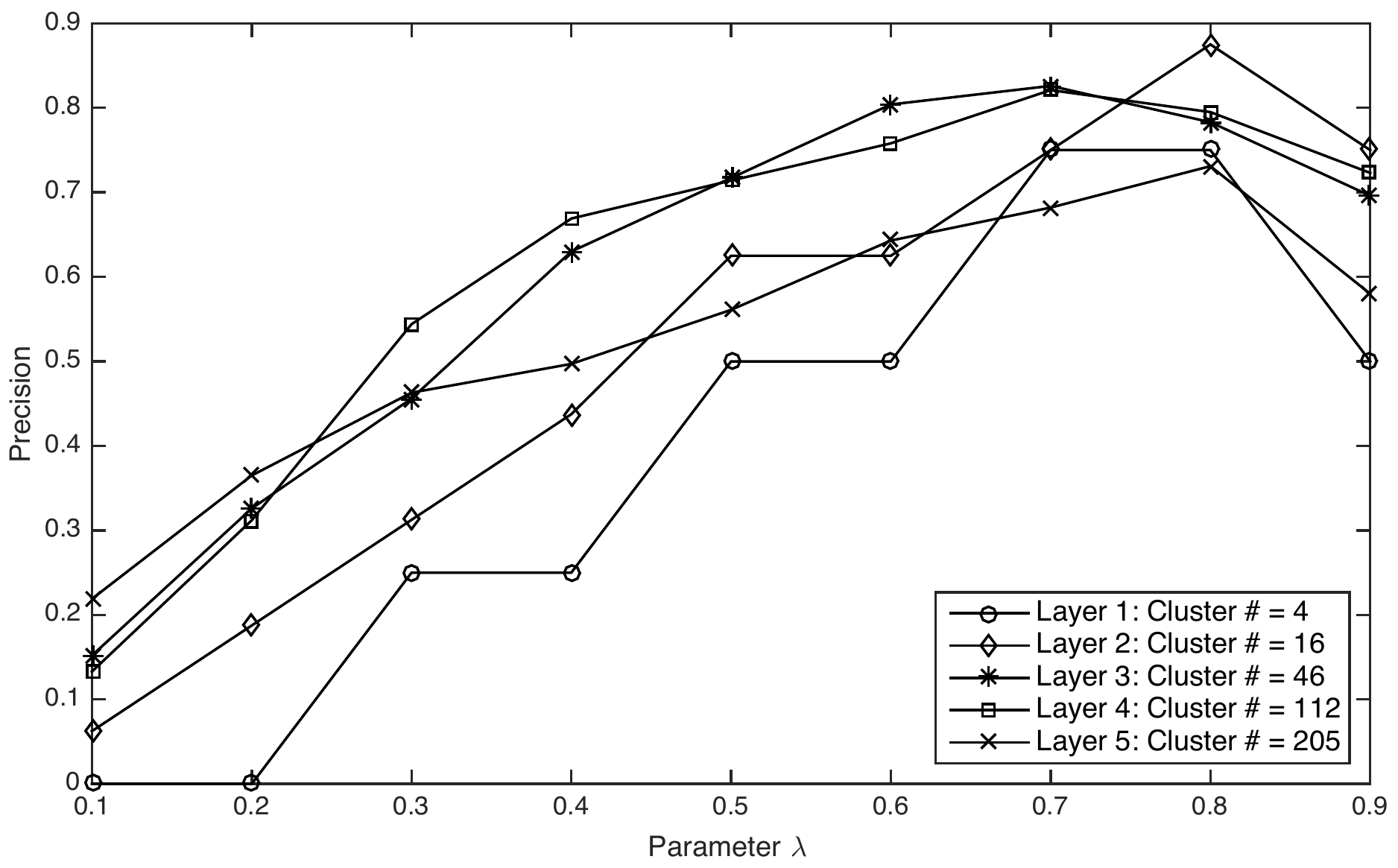}
\caption{Naming Performance Comparison in Terms of $\lambda$}
\label{naming}
\end{figure}

In order to evaluate the type naming performance, we ask 3 human annotators to determine whether the type label matches the reference class label on the English news data set. We tune the parameter $\lambda$ which is used in Section~\ref{sec:typeNaming} to find the optimal threshold for type naming, which is shown in Figure~\ref{naming}. From the comparison results, when $\lambda$ is 0.8, the naming performance is close to $90\%$.

\section{Conclusions and Future Work}

In this work, we take a fresh look at the fine-grained entity typing problem and, for the first time, propose an unsupervised framework, which incorporates entity general semantics, specific contexts and domain specific knowledge to discover the fine-grained types. This framework takes the human out of the loop and requires no annotated data or predefined types. Without the needs of language-specific features and resources, this framework can be easily adapted to other domains, genres and languages. We also incorporate a domain- and language-independent unsupervised entity linking system to improve the clustering performance and discover corpus-customized domain-specific fine-grained typing schema. Our framework achieves comparable performance to state-of-the-art entity typing systems trained from a large amount of labeled data. In the future, we will explore joint fine-grained entity linking and typing mentions for mutual enhancement.  

%




\bibliographystyle{abbrv}
\bibliography{sigproc}  

\end{document}